\documentclass[runningheads]{llncs}

\usepackage{graphicx}

\usepackage[utf8]{inputenc} 
\usepackage{enumitem}
\usepackage{verbatim}
\usepackage{booktabs}       

\usepackage[colorlinks, citecolor=black, linkcolor=black, urlcolor=blue]{hyperref}
\begin{document}

\title{Visualizing Convolutional Networks for \protect\\ MRI-based Diagnosis of Alzheimer's Disease}

\titlerunning{Visualizing CNNs for MRI-based Diagnosis of Alzheimer's Disease}

\author{Johannes Rieke\inst{1,2} \and
Fabian Eitel\inst{1} \and
Martin Weygandt\inst{1} \and
John-Dylan Haynes\inst{1} \and
Kerstin Ritter\inst{1} \and
for the Alzheimer’s Disease Neuroimaging Initiative\thanks{Data used in preparation of this article were obtained from the Alzheimer’s Disease Neuroimaging Initiative (ADNI) database (adni.loni.usc.edu). As such, the investigators within the ADNI contributed to the design and implementation of ADNI and/or provided data but did not participate in analysis or writing of this report. A complete listing of ADNI investigators can be found at: \url{http://adni.loni.usc.edu/wp-content/uploads/how_to_apply/ADNI_Acknowledgement_List.pdf}}}
\authorrunning{J. Rieke et al.}

\institute{Charit\'e – Universit\"atsmedizin Berlin, corporate member of Freie
Universit\"at Berlin, Humboldt-Universit\"at zu Berlin, and Berlin Institute of
Health (BIH); Bernstein Center for Computational Neuroscience, Berlin
Center for Advanced Neuroimaging, Department of Neurology, and Excellence
Cluster NeuroCure\and 
Technical University Berlin\\
\email{johannes.rieke@gmail.com}}

\maketitle              

\begin{abstract}
Visualizing and interpreting convolutional neural networks (CNNs) is an important task to increase trust in automatic medical decision making systems. In this study, we train a 3D CNN to detect Alzheimer’s disease based on structural MRI scans of the brain. Then, we apply four different gradient-based and occlusion-based visualization methods that explain the network’s classification decisions by highlighting relevant areas in the input image. We compare the methods qualitatively and quantitatively. We find that all four methods focus on brain regions known to be involved in Alzheimer's disease, such as inferior and middle temporal gyrus. While the occlusion-based methods focus more on specific regions, the gradient-based methods pick up distributed relevance patterns. Additionally, we find that the distribution of relevance varies across patients, with some having a stronger focus on the temporal lobe, whereas for others more cortical areas are relevant. In summary, we show that applying different visualization methods is important to understand the decisions of a CNN, a step that is crucial to increase clinical impact and trust in computer-based decision support systems.
\end{abstract}

\section{Introduction}
Alzheimer’s disease (AD) is the main cause of dementia in the elderly. It is symptomatically characterized by loss of memory and other intellectual abilities to such an extent that it affects daily life. Long before memory problems occur, microscopic changes related to cell death take place and slowly progress over time. Radiologically, neurodegeneration is the hallmark of AD, starting in the temporal lobe and then spreading all over the brain. However, since all brains from elderly people are affected by atrophy, it is a difficult task (even for experienced radiologists) to discriminate normal age-related atrophy from AD-mediated atrophy.

In this context, machine learning models provide great potential to capture even slight tissue alterations. State-of-the-art models for image classification are convolutional neural networks (CNNs), which have recently been applied to medical imaging data for various use cases \cite{Litjens2017}, including AD detection. The key idea behind CNNs is inspired by the mechanism of receptive fields in the primate visual cortex: Local convolutional filters and pooling operations are applied successively to extract regional information. In contrast to traditional machine learning-based approaches, CNNs do not rely on hand-crafted features but find meaningful representations of the input data during training.

Although CNNs deliver good classification results, they are difficult to visualize and interpret. In medical decision making, however, it is critical to explain the behavior of a machine learning model and let medical experts verify the diagnosis. A number of visualization methods have been suggested that highlight regions in an input image with strong influence on the classification decision \cite{Simonyan2014,Springenberg2015,Zeiler2014,Yang2018VisualClassification}. Such heatmaps constitute the basis for understanding and interpreting machine learning models, optimally together with clinicians.

In this work, we compare four visualization methods (sensitivity analysis, guided backpropagation, occlusion and brain area occlusion) on a 3D CNN, which was trained to classify structural MRI scans of the brain into AD patients and normal elderly controls (NCs).

\section{Related Work}
\subsection{Alzheimer Classification}
A number of machine learning models have been applied to Alzheimer detection. Some use traditional approaches with hand-crafted features, while most recent papers employ deep convolutional networks. For an overview, we refer the reader to Table 1 in Khvostikov et al. \cite{Khvostikov2018}. We identified three studies that use a model and training procedure similar to ours (i.e. 3D CNN, full-brain structural MRI scans, AD/NC classification) \cite{Payan2015,Hosseini-Asl2018,Korolev2017}: In contrast to our study, Payan et al. \cite{Payan2015} and Hosseini-Asl et al. \cite{Hosseini-Asl2018} pretrain their convolutional layers with an unsupervised autoencoder. Korolev et al. \cite{Korolev2017} train from scratch, but use more complex networks. The CNN architecture in our study is partly inspired by a model in Khvostikov et al. \cite{Khvostikov2018}, even though they only train on images of the hippocampus. 

\subsection{Visualization Methods}
A range of visualization methods for CNNs have recently been developed. While some methods aim to find prototypes for classes, we only use methods that explain the CNN decision for a specific sample \cite{Simonyan2014,Springenberg2015,Zeiler2014} (see methodological details below). We found two studies that apply visualization methods to AD classification in a similar way as we do: Korolev et al. \cite{Korolev2017} employ the occlusion method on a deep CNN. While they show similar results like ours (focus on hippocampus and ventricles), they do not compare different visualization methods or analyze the relevance distribution in detail. Yang et al. \cite{Yang2018VisualClassification} use a segmentation-based occlusion (similar to our brain area occlusion), but reach inconclusive results. To the best of our knowledge, this is the first study that comprehensively compares different visualization methods on CNNs for AD detection.

\section{Methods}
\subsection{Data}
Data used in the preparation of this article were obtained from the Alzheimer’s Disease Neuroimaging Initiative (ADNI) database (\url{adni.loni.usc.edu}). The ADNI was launched in 2003 as a public-private partnership, led by Principal Investigator Michael W. Weiner, MD. The primary goal of ADNI has been to test whether serial magnetic resonance imaging (MRI), positron emission tomography (PET), other biological markers, and clinical and neuropsychological assessment can be combined to measure the progression of mild cognitive impairment (MCI) and early Alzheimer’s disease (AD).

For this study we used structural MRI data of patients with Alzheimer's disease (AD) and normal controls (NC) from phase 1 of ADNI who were included in the "MRI collection - Standardized 1.5T List - Annual 2 year". For each subject, this data collection offers structural MRI scans of the full brain for up to three time points (screening, 12 and 24 months; sometimes multiple scans per visit). We excluded scans with mild cognitive disorder (MCI) and two scans for which our preprocessing pipeline failed. In total, our dataset comprises 969 individual scans (475 AD, 494 NC) from 344 subjects (193 AD, 151 NC). 

All scans were acquired with 1.5 Tesla scanners at various sites and had  undergone gradient non-linearity, intensity inhomogeneity and phantom-based distortion correction. We downloaded T1-weighted MPRAGE scans and non-linearly registered all images to a 1 mm isotropic ICBM template using ANTs (\url{http://stnava.github.io/ANTs/}), resulting in volumes of $193 \times 229 \times 193$.  

For training, we split this dataset using 5-fold cross validation. The split is performed on the level of patients to prevent the network from seeing images of the same patient during training and testing. For the visualization methods, we used a fixed split with 30 AD and 30 NC patients in the test set. 

\subsection{Model}
Our model consists of four convolutional layers (with filter size $3 \times 3 \times 3$ and 8/16/32/64 feature maps) and two fully-connected layers (128/64 neurons; this architecture is inspired by a model in Khvostikov et al. \cite{Khvostikov2018}). We apply batch normalization and pooling after each convolution and dropout of 0.8 before the first fully-connected layer. The network has two output neurons with softmax activation. We train with cross-entropy loss and the Adam optimizer (learning rate 0.0001, batch size 5) for 20 epochs. Before feeding the brain scans to the network, we remove the skull and normalize each voxel to have mean 0 and standard deviation 1 across the training set. 

\subsection{Visualization Methods}
In this section, we briefly review the four visualization methods we used in this study (see also the review by Montavon et al. \cite{Montavon2018}). All of these methods produce a heatmap over the input image, which indicates the relevance of image pixels for the classification decision. PyTorch implementations of all visualization methods will be made available at \url{http://github.com/jrieke/cnn-interpretability}.

\subsubsection{Sensitivity Analysis (Backpropagation) \cite{Simonyan2014}}
The gradient of the network’s output probability w.r.t. the input image is calculated. For a given image pixel, this gradient describes how much the output probability changes when the pixel value changes. In neural networks, the gradient can be easily computed via the backpropagation algorithm, which is used for training.  As relevance score, we take the absolute value of the gradient. 

\subsubsection{Guided Backpropagation \cite{Springenberg2015}} This method is a modified version of sensitivity analysis, in which the negative gradients are set to 0 at ReLU layers during the backward pass. This is equivalent to a combination of Backpropagation and Deconvolution and leads to more focused heatmaps. As above, we take the absolute value of the gradient as the relevance score. 

\subsubsection{Occlusion \cite{Zeiler2014}} A part of the image is occluded with a black or gray patch and the network output is recalculated. If the probability for the target class decreases compared to the original image, this image region is considered to be relevant. To get a relevance heatmap, we slide the patch across the image and plot the difference between unoccluded and occluded probability (for AD or NC). We use a patch of size $40 \times 40 \times 40$ with value 0. 

\subsubsection{Brain Area Occlusion} This method is a modification of occlusion, in which we occlude an entire brain area based on the Automated Anatomical Labeling atlas (AAL, \url{http://www.gin.cnrs.fr/en/tools/aal-aal2/}). This method was inspired by a segmentation-based visualization in Yang et al. \cite{Yang2018VisualClassification}.As for occlusion, we report the difference between unoccluded and occluded probability (for AD or NC). 

\begin{figure}[hp]
  \vspace*{0cm}
  \centerline{\includegraphics[width=1.08\textwidth]{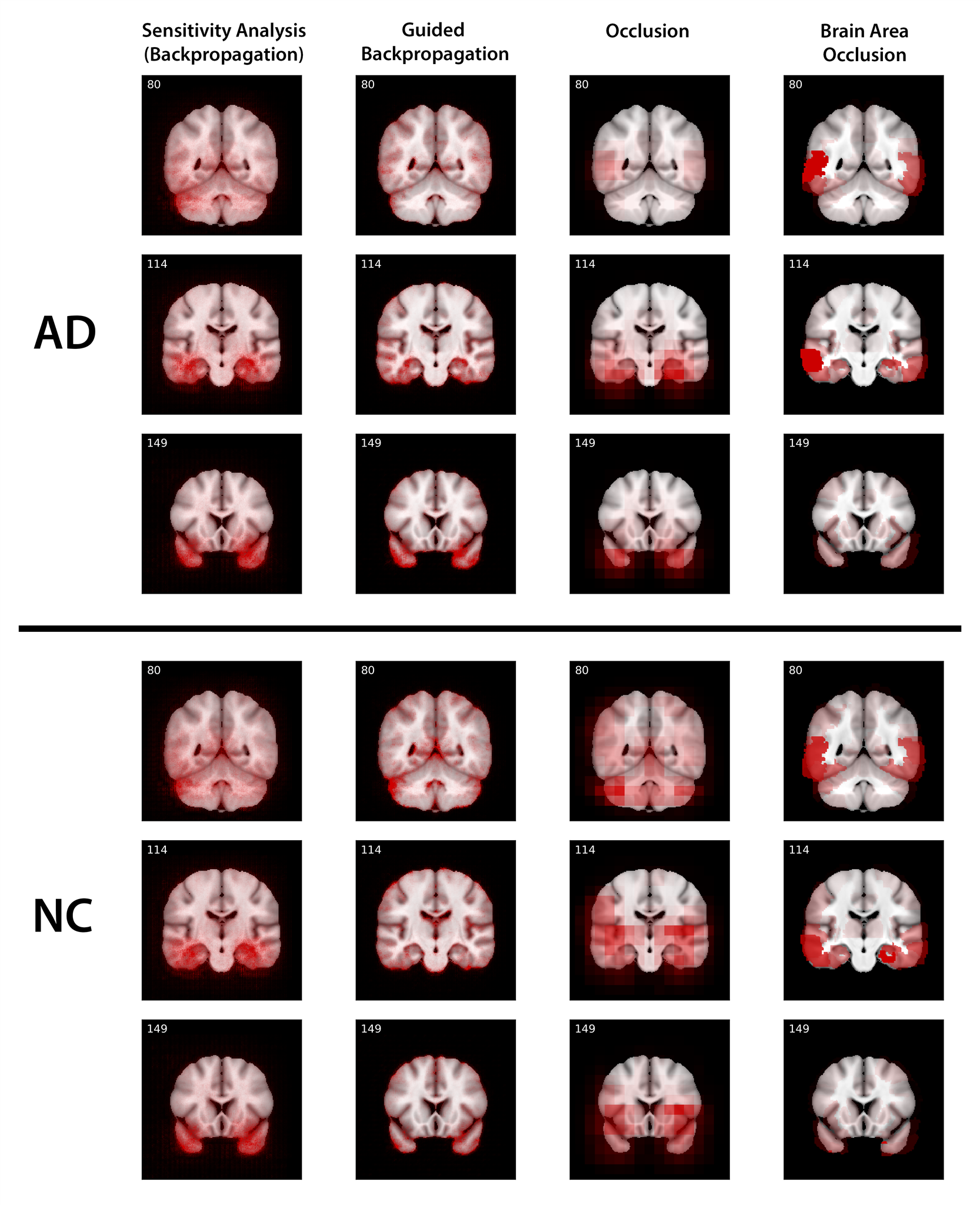}}
  \vspace*{0cm}
  \caption{Relevance heatmaps for all visualization methods, averaged over AD (top) and NC (bottom) samples in the test set. Red indicates relevance, i.e. a red area was important for the network's classification decision. Numbers indicate slice positions (out of 229 coronal slices).}
  \label{fig-heatmaps} 
\end{figure}

\section{Results}
\subsection{Classification}
Using 5-fold cross-validation, our network achieves a classification accuracy of $0.77 \pm 0.06$ and ROC AUC of $0.78 \pm 0.04$ (both mean $\pm$ standard deviation). This is comparable to recent studies for other convolutional networks \cite{Payan2015,Hosseini-Asl2018,Korolev2017}. For example, Korolev et al. \cite{Korolev2017}, who use a similar model and training procedure, achieve a similar accuracy of $0.79 \pm 0.08$, but with a better ROC AUC of $0.88 \pm 0.08$. Please note that our focus was on the different visualization methods and not on optimizing the network. 

\begin{table}[b]
\centering
\small
\caption{Most relevant brain areas per visualization method, averaged over AD (top) and NC (bottom) samples in the test set. Values in brackets give fraction of summed relevance in this brain area, divided by the summed relevance in the whole brain.}
\label{table-summed-relevance}
\centerline{
\setlength{\tabcolsep}{0.5em} 
\begin{tabular}{lcccc}
\hline
\multicolumn{1}{|c|}{} & \multicolumn{1}{c|}{\begin{tabular}[c]{@{}c@{}}Sensitivity Analysis\\ (Backpropagation)\end{tabular}} & \multicolumn{1}{c|}{\begin{tabular}[c]{@{}c@{}}Guided \\ Backpropagation\end{tabular}} & \multicolumn{1}{c|}{Occlusion} & \multicolumn{1}{c|}{\begin{tabular}[c]{@{}c@{}}Brain Area\\ Occlusion\end{tabular}} \\ \hline
\multicolumn{1}{|l|}{AD} & \multicolumn{1}{c|}{\begin{tabular}[c]{@{}c@{}}TemporalMid (6.1 \%)\\ TemporalInf (5.9 \%)\\ Fusiform (4.6 \%)\\ CerebelumCrus1 (3.8 \%)\end{tabular}} & \multicolumn{1}{c|}{\begin{tabular}[c]{@{}c@{}}TemporalMid (7.0 \%)\\ TemporalInf (5.7 \%)\\ FrontalMid (4.2 \%)\\ Fusiform (3.9 \%)\end{tabular}} & \multicolumn{1}{c|}{\begin{tabular}[c]{@{}c@{}}TemporalMid (12.1 \%)\\ TemporalInf (9.2 \%)\\ Fusiform (6.2 \%)\\ ParaHippocampal (5.4 \%)\end{tabular}} & \multicolumn{1}{c|}{\begin{tabular}[c]{@{}c@{}}TemporalMid (29.7 \%)\\ TemporalInf (14.8 \%)\\ TemporalSup (4.4 \%)\\ Hippocampus (4.1 \%)\end{tabular}} \\ \hline
\multicolumn{1}{|l|}{NC} & \multicolumn{1}{c|}{\begin{tabular}[c]{@{}c@{}}TemporalMid (6.1 \%)\\ TemporalInf (5.8 \%)\\ Fusiform (4.5 \%)\\ CerebelumCrus1 (3.8 \%)\end{tabular}} & \multicolumn{1}{c|}{\begin{tabular}[c]{@{}c@{}}CerebelumCrus1 (4.6 \%)\\ TemporalMid (4.5 \%)\\ TemporalInf (4.5 \%)\\ FrontalMid (4.1 \%)\end{tabular}} & \multicolumn{1}{c|}{\begin{tabular}[c]{@{}c@{}}TemporalMid (6.2 \%)\\ TemporalSup (4.9 \%)\\ CerebelumCrus1 (4.9 \%)\\ Insula (4.7 \%)\end{tabular}} & \multicolumn{1}{c|}{\begin{tabular}[c]{@{}c@{}}TemporalMid (20.4 \%)\\ TemporalInf (12.8 \%)\\ Fusiform (7.2 \%)\\ TemporalSup (6.2 \%)\end{tabular}} \\ \hline
 & \multicolumn{1}{l}{} & \multicolumn{1}{l}{} & \multicolumn{1}{l}{} & \multicolumn{1}{l}{}
\end{tabular}
}
\end{table}

\subsection{Relevant brain areas}
Fig. \ref{fig-heatmaps} shows relevance heatmaps for all visualizations methods, averaged over AD and NC samples in the test set. Since there is no ground truth available for such heatmaps, we validate our results by focusing on specific brain areas that were associated with AD in the medical literature. We identified the most relevant brain areas for each visualization method by summing the relevance in each area (according to the AAL atlas). Table \ref{table-summed-relevance} lists the four most relevant brain areas for each method, again averaged over AD and NC samples. 

For both AD and NC patients, we can see that the main focus of the network is on the temporal lobe, especially its medial part. This brain area, containing the hippocampus and other structures associated with memory, has been empirically linked to AD \cite{Frisoni2010}. The hippocampus itself is usually one of the earliest areas affected by AD \cite{Mu2011}. In our experiments, we observe some relevance on the hippocampus, but usually the whole area around it is crucial for the network's decision. This may be explained by the fact that our samples contain only advanced forms of the disease. 

In addition to temporal regions, we observe some relevance attributed to other areas across the brain (especially in the gradient-based visualization methods). We find that the distribution of relevance varies between patients: Some brains have strong relevance in the temporal lobe, while in others, the cortex plays a crucial role. 

Lastly, we note that the heatmaps for AD and NC samples are quite similar. This makes sense, given that the network should focus on the same regions to detect presence or absence of the disease. Some differences between AD and NC can be found for the occlusion method. We speculate that this might be an artifact of our specific setting (the network might confuse the occlusion patch with brain atrophy, increasing the probability for AD in some brain areas). 

\begin{table}[b]
\centering
\caption{Euclidean distance between relevance heatmaps (averaged over all AD / NC samples in the test set) in $10^{-4}$.}
\label{table-heatmaps-distance}
\begin{tabular}{|c|c|c|c|c|}
\hline
\setlength{\tabcolsep}{0.5em} 
\renewcommand{\arraystretch}{1.2}
 & \begin{tabular}[c]{@{}c@{}}Sensitivity Analysis\\ (Backpropagation)\end{tabular} & \begin{tabular}[c]{@{}c@{}}Guided\\ Backpropagation\end{tabular} & Occlusion & \begin{tabular}[c]{@{}c@{}}Brain Area\\ Occlusion\end{tabular} \\ \hline
\begin{tabular}[c]{@{}c@{}}Sensitivity Analysis\\ (Backpropagation)\end{tabular} & 0.00 / 0.00 & 4.09 / 4.36 & 5.15 / 4.09 & 11.48 / 9.04 \\ \hline
\begin{tabular}[c]{@{}c@{}}Guided\\ Backpropagation\end{tabular} & 4.09 / 4.36 & 0.00 / 0.00 & 6.47 / 5.83 & 11.36 / 9.80 \\ \hline
Occlusion & 5.15 / 4.09 & 6.47 / 5.83 & 0.00 / 0.00 & 11.16 / 9.66 \\ \hline
\begin{tabular}[c]{@{}c@{}}Brain Area\\ Occlusion\end{tabular} & 11.48 / 9.04 & 11.36 / 9.80 & 11.16 / 9.66 & 0.00 / 0.00 \\ \hline
\end{tabular}
\end{table}

\subsection{Differences between visualization methods}
Although all visualization methods focus on similar brain areas, we can spot some differences: Occlusion and brain area occlusion are more focused on specific regions, while relevance in the gradient-based methods seems more distributed. Obviously, the occlusion-based approaches cannot deal with large areas of distributed relevance (e.g. in the cortex), because these areas will never be covered up completely by the occlusion patch. Therefore, we recommend to apply gradient-based instead of occlusion-based visualization methods for use cases where the relevance is presumably distributed across the input image. Moreover, we find that brain area occlusion is indeed a very natural approach for our context, but it suffers from the fact that only one brain region is covered up at a time. In our case, this leads to very high relevance for the temporal lobe, but hardly any relevance for other brain structures. 

To compare the visualization methods quantitatively, we computed Euclidean distances between all average heatmaps ($\sqrt{\sum_i{(A_i - B_i)^2}}$, where $A$ and $B$ are the average heatmaps of two distinct methods and $i$ is the voxel location). The distances are shown in Table \ref{table-heatmaps-distance}. In accordance with the visual impression, we find that the gradient-based methods are relatively similar to each other (i.e. low Euclidean distance). The only method that deviates strongly from all other methods is brain area occlusion, which (as stated above) only attributes relevance to a few image regions.

\section{Conclusion}
In this study, we trained a 3D CNN for Alzheimer classification and applied various visualization methods. We show that our CNN indeed focuses on brain regions associated with AD, in particular the medial temporal lobe. This is consistent across all four visualization methods. Interestingly, the distribution of relevance varies between patients, with some having a stronger focus on the temporal lobe, whereas for others more cortical areas were involved. We hope that explaining classifier decisions in this way can pave the way for machine learning models in critical areas like medicine and will increase trust in computer-based decision support systems. Our results also show that the visualization methods differ in their explanations. Therefore, we strongly recommend to compare available visualization methods for a specific application area and not “blindly” trust the results of one method. 

For future research, we identified three main areas: First, other visualization methods \cite{Montavon2018} could be implemented and compared to our results. Second, future studies might apply our workflow to preconditions of Alzheimer's disease, i.e. mild cognitive impairment, and measures of clinical disability. Third, it would be interesting to produce some form of ground truth for the relevance heatmaps, e.g. by implementing simulation models that control for the level of separability or location of differences.

\section*{Acknowledgements}
\small
\begin{sloppypar}
This research was funded by the Deutsche Forschungsgemeinschaft (DFG, 389563835) and a Rahel-Hirsch scholarship from Charité - Universitätsmedizin Berlin. The Titan Xp used for this research was donated by the NVIDIA Corporation.

Data collection and sharing for this project was funded by the Alzheimer's Disease
Neuroimaging Initiative (ADNI) (National Institutes of Health Grant U01 AG024904) and
DOD ADNI (Department of Defense award number W81XWH-12-2-0012). ADNI is funded
by the National Institute on Aging, the National Institute of Biomedical Imaging and
Bioengineering, and through generous contributions from the following: AbbVie, Alzheimer’s
Association; Alzheimer’s Drug Discovery Foundation; Araclon Biotech; BioClinica, Inc.;
Biogen; Bristol-Myers Squibb Company; CereSpir, Inc.; Cogstate; Eisai Inc.; Elan
Pharmaceuticals, Inc.; Eli Lilly and Company; EuroImmun; F. Hoffmann-La Roche Ltd and
its affiliated company Genentech, Inc.; Fujirebio; GE Healthcare; IXICO Ltd.; Janssen
Alzheimer Immunotherapy Research \& Development, LLC.; Johnson \& Johnson
Pharmaceutical Research \& Development LLC.; Lumosity; Lundbeck; Merck \& Co., Inc.;
Meso Scale Diagnostics, LLC.; NeuroRx Research; Neurotrack Technologies; Novartis
Pharmaceuticals Corporation; Pfizer Inc.; Piramal Imaging; Servier; Takeda Pharmaceutical
Company; and Transition Therapeutics. The Canadian Institutes of Health Research is
providing funds to support ADNI clinical sites in Canada. Private sector contributions are
facilitated by the Foundation for the National Institutes of Health (\url{www.fnih.org}). The grantee
organization is the Northern California Institute for Research and Education, and the study is
coordinated by the Alzheimer’s Therapeutic Research Institute at the University of Southern
California. ADNI data are disseminated by the Laboratory for Neuro Imaging at the
University of Southern California. 
\end{sloppypar}

%
%
\bibliographystyle{splncs04}
\bibliography{Mendeley}

\end{document}